\documentclass[lettersize,journal]{IEEEtran}
\usepackage{amsmath,amsfonts}
\usepackage{algorithmic}
\usepackage{algorithm}
\usepackage{array}
\usepackage{xcolor}
\usepackage[caption=false,font=normalsize,labelfont=sf,textfont=sf]{subfig}
\usepackage{textcomp}
\usepackage{stfloats}
\usepackage{url}
\usepackage{verbatim}
\usepackage{graphicx}
\usepackage{cite}
\usepackage{soul}
\usepackage{tablefootnote}

\usepackage{lmodern}
\usepackage{makecell}
\usepackage{float}
\usepackage{longtable}
\usepackage{multirow}
\usepackage{iftex}
\usepackage{ragged2e} 
\usepackage{hyperref}
\usepackage{fancybox}  

\begin{document}

\title{A Cascaded Unsupervised–Supervised NLP Pipeline for Detecting Accusatory Language in Public Procurement}
\author{
Bryan~Torres,~\IEEEmembership{Member,~IEEE},
Daniel Riofrío,~\IEEEmembership{Member,~IEEE},
Jos\'e~Vega-S\'anchez,~\IEEEmembership{Senior~Member,~IEEE},
Nathaly~Orozco~Garz\'on,~\IEEEmembership{Senior~Member,~IEEE},
Carla Parra,
Karen Rosero,~\IEEEmembership{Member,~IEEE},
and Felipe Grijalva,~\IEEEmembership{Senior~Member,~IEEE}%
\thanks{Manuscript received Month XX, 2026; revised Month XX. The review of this paper was coordinated by XXXX. This work was supported in part by the Universidad San Francisco de Quito through the Poli-Grants Program under Grants 41989 and 41990, and by the Universidad de Las Américas under Project 578.A.XVII.25. Corresponding author: Felipe~Grijalva (E-mail: fgrijalva@usfq.edu.ec)}
\thanks{B.~Torres, D. Riofrío, J.~Vega-S\'anchez and F. Grijalva are with the Colegio de Ciencias e Ingenier\'ias  ``El Polit\'ecnico'', Universidad San Francisco de Quito (USFQ), Diego de Robles S/N, Quito 170157 (e-mail: btorres@estud.usfq.edu.ec; \{driofrioa,dvega,fgrijalva\}@usfq.edu.ec).}%
\thanks{N.~Orozco~Garz\'on is with the Faculty of Engineering and Applied Sciences, Telecommunications Engineering, ETEL Research Group, Universidad de Las Américas (UDLA), 170503 Quito, Ecuador (e-mail: nathaly.orozco@udla.edu.ec).}%
\thanks{C.~Parra is with the Departamento de Estudios Organizacionales y Desarrollo Humano, Escuela Politécnica Nacional, Quito 170525, Ecuador (e-mail: carla.parra@epn.edu.ec).}%
\thanks{K. Rosero is with the Language Technologies Institute, Carnegie Mellon University, Pittsburgh, PA, USA (e-mail: kroseroj@andrew.cmu.edu).}%
}
\maketitle

\begin{abstract}
Public procurement involves the allocation of substantial financial resources; therefore, continuous oversight through audits, controls, and monitoring mechanisms is essential. However, stakeholder comments and publicly available government data are often underutilized, despite their potential to reveal procedural irregularities. To address this gap, this paper analyzes metadata from Ecuador’s \textit{Sistema Oficial de Contratación Pública} (SOCE, Official Public Procurement System), with particular emphasis on participant comments generated during the pre-contractual phase. We propose a hybrid modeling framework that integrates unsupervised clustering and supervised classification within a natural language processing (NLP) pipeline to uncover latent patterns and detect potentially irregular procurement processes. Semantic embeddings are generated using Word2Vec, LLaMA, and RoBERTa, followed by Gaussian Mixture Models (GMMs) for unsupervised clustering. A supervised classification stage is then applied to identify accusatory or whistleblowing-style comments.
Experimental results show that the combination of domain-trained Word2Vec embeddings, GMM-based clustering, and a Random Forest classifier achieves high precision and recall, even under severe class imbalance. These findings demonstrate that lightweight, domain-adapted NLP architectures can effectively support risk identification and enhance transparency in public procurement systems without requiring large-scale computational infrastructure.
\textit{\textbf{Reproducible Research:}} The simulation code is available at \href{https://github.com/Espejin/USFQ-MSDS-Accusatory-Comments-Classification}{\textbf{$<$Kapak NLP Pipeline$>$}}.
\end{abstract}


\section{Introduction}
\IEEEPARstart{C}{orruption} is a pervasive phenomenon that persists across societies and historical periods. Ecuador is no exception: in 2024, Transparency International reported a Corruption Perceptions Index (CPI) score of 32 out of 100 for the country, where lower values indicate higher perceived corruption \cite{transparency2024}. In parallel, public procurement processes reached 7.996 billion USD, representing 24\% of the national General State Budget \cite{SERCOP2024}.
In this context, Ecuador has introduced multiple policies to mitigate corruption in public procurement, motivated by evidence of biased procedures and limited transparency in both judicial actions and procurement processes. Notably, the National Anti-Corruption Strategy, planned through 2025, aims to promote transparency and open data across government operations. Earlier initiatives, such as Ecuador's first Open Government Action Plan, emphasized transparency, access to public information, and citizen participation as key elements for strengthening government accountability \cite{OGPEcuador}. Although platforms such as the \textit{Sistema Oficial de Contratación Pública} (SOCE) provide extensive access to procurement data, effective use of this information requires technical expertise. As a result, the data remains difficult to interpret for both oversight institutions and the general public, limiting its practical impact \cite{ulloa2024kapakdata}.

In response to this challenge, the Universidad San Francisco de Quito (USFQ) developed the Kapak project, which aims to generate data-driven indicators for identifying potential corruption risks in public procurement. Kapak focuses on two procurement modalities widely used in Ecuador: the Reverse Electronic Auction (SIE), where suppliers progressively reduce their bids, and the Specific Business Line (GEN), which groups recurrent procurement categories. The project relies on publicly available data from Ecuador’s \textit{Servicio Nacional de Contratación Pública} (SERCOP), accessed through the SOCE, to produce statistical analyses across all stages of the procurement cycle \cite{kapak_metodologia}. Specifically, Kapak employs automated web crawlers to continuously extract procurement data in its native SOCE format, enabling the preservation of historical records and reducing the risk of data loss due to irregular modifications. The collected data is stored in a dedicated database, where it is cleaned and processed to support targeted analyses. The final outcome is a set of composite corruption risk indicators that facilitate the identification of high-risk procurement processes and support more efficient auditing and oversight efforts \cite{kapak_metodologia}.

In recent years, deep learning-based natural language analysis has gained significant momentum, demonstrating broad applicability across a wide range of domains \cite{minaee2021deep}. Nevertheless, natural language processing (NLP) applications have been developed for decades, supported by a wide range of frameworks addressing language-related tasks. One of the core problems in NLP is text classification, where textual inputs are mapped to predefined categories, a task extensively studied in both academia and industry. A common strategy relies on keyword-based methods for sentiment analysis or topic detection \cite{cambria2014senticnet}. For instance, sentiment prediction in social media has been achieved using keyword filtering combined with classifiers such as Complement Naive Bayes \cite{sentiment_analysis_IMDEA_2013}. However, keyword-based approaches often struggle with ambiguity, semantic shifts, and contextual nuances, limitations that become more severe in unsupervised learning scenarios. In contrast, structured textual data—such as scientific publications—has been effectively analyzed using clustering techniques and low- to medium-complexity vectorization methods, including term frequency (TF) and term frequency–inverse document frequency (TF-IDF) \cite{paper_cluster_2022}. The syntactic regularity and well-defined structure of such documents facilitate preprocessing and clustering, often allowing sections to be analyzed independently \cite{yau2014_pap_clustering}. Conversely, user-generated text, such as public procurement comments, is typically noisy and unstructured, increasing preprocessing complexity and requiring more robust text normalization strategies.

{While applications of modern Large Language Models (LLMs) in the field of corruption detection remain scarce, recent years have seen the emergence of methods for detecting accusatory phrases in irregular processes. These approaches employ prompting techniques, such as greedy and beam search, utilizing state-of-the-art LLMs like GPT-4o-mini, LLaMA 3.1, and Phi 3.5 mini. Although these applications have yielded promising results—with AUC values ranging from 0.88 to 0.97. The experiments themselves reveal a critical challenge regarding class imbalance; specifically, these models exhibit high recall but very low precision} \cite{bn_kap_prompt_clas}. {Additionally, they encounter the typical drawbacks associated with LLM deployment: high monetary costs per token, intensive computational resource requirements, and excessive processing time.}

Building upon the previous analysis, the application of machine learning techniques for detecting corruption remains relatively underexplored, largely due to the scarcity of large, reliable, and high-quality labeled datasets. As a result, government-issued documents are commonly used as primary data sources, despite often requiring segmentation into relevant sections to extract meaningful information \cite{prediction_corruption_2019}. Although their institutional origin provides a certain level of structure and formality, data quality and transparency vary significantly across institutions. In this context, the Organisation for Economic Co-operation and Development (OECD) reported that among 39 surveyed countries, 74\% do not actively use artificial intelligence (AI) in anti-corruption or integrity processes, despite ongoing exploratory efforts \cite{Ugale2024}. This finding highlights the limited availability of actionable data and AI-driven practices for analyzing public processes, a challenge that is particularly pronounced in Latin American and African countries.

Several studies have attempted to identify objective and quantifiable variables to analyze corruption-related phenomena using machine learning and artificial intelligence techniques \cite{aldana2022machinelearningmodelidentify}. While promising results have been reported, these approaches remain constrained by the limited reliability and trustworthiness of data obtained from governmental sources. Moreover, the scarcity of labeled corruption data has led some studies to rely on proxy or complementary indices as target variables, effectively reformulating classification tasks as regression problems. However, the instability of these surrogate models often produces unreliable outcomes, particularly under conditions of low data transparency and severe class imbalance~\cite{prediction_corruption_2019}.

To address the limitations of institutionally generated data, a promising alternative is to leverage third-party content, such as comments submitted by procurement participants. This user-generated feedback reflects direct experiences with procurement processes and can reveal irregularities not captured in official reporting, particularly in low-transparency settings. Building on this insight, the present study is conducted within the framework of the Kapak initiative, leveraging its infrastructure for large-scale data collection and analysis to enhance corruption risk indicators in public procurement.  This work focuses on detecting accusatory or whistleblowing-style comments expressed during the pre-contractual phase using NLP techniques tailored to noisy, user-generated text. We propose a cascaded hybrid learning strategy that integrates unsupervised clustering and supervised classification. Domain-specific embeddings generated using Word2Vec, RoBERTa, and LLaMA are used to cluster unlabeled data via Gaussian Mixture Models (GMMs), followed by supervised classification to identify accusatory content. In summary, the main contributions of this paper are as follows:
\begin{itemize}
    \item A hybrid architecture combining unsupervised clustering and supervised classification for detecting accusatory comments in Ecuador’s public procurement system.
    \item A comparative evaluation of embedding strategies (LLaMA, RoBERTa, and Word2Vec), showing that a customized Word2Vec model performs best on noisy, Spanish-language, user-generated text.
    \item An effective GMM-based clustering strategy, validated using Silhouette Score and keyword-based heuristics to identify clusters enriched with accusatory content.
    \item A lightweight Random Forest classifier achieving high performance, with a precision of 0.84 and recall of 0.91 under severe class imbalance.
    \item A computationally efficient, semi-supervised pipeline that can be deployed without large-scale infrastructure for real-time auditing and large-scale monitoring.
\end{itemize}
The remainder of this paper is organized as follows. Section~\ref{sec:materials_methods} describes the datasets, preprocessing steps, and the proposed NLP pipeline. 
Section~\ref{sec:results} presents and discusses the experimental results. 
Section~\ref{sec:generalizability} analyzes cross-country generalizability and data availability in Latin America. 
Section~\ref{sec:conclu} concludes the paper and outlines future research directions.

\section{Materials and Methods}\label{sec:materials_methods}
\begin{figure*}[t!]
    \centering
    \includegraphics[scale=0.36]{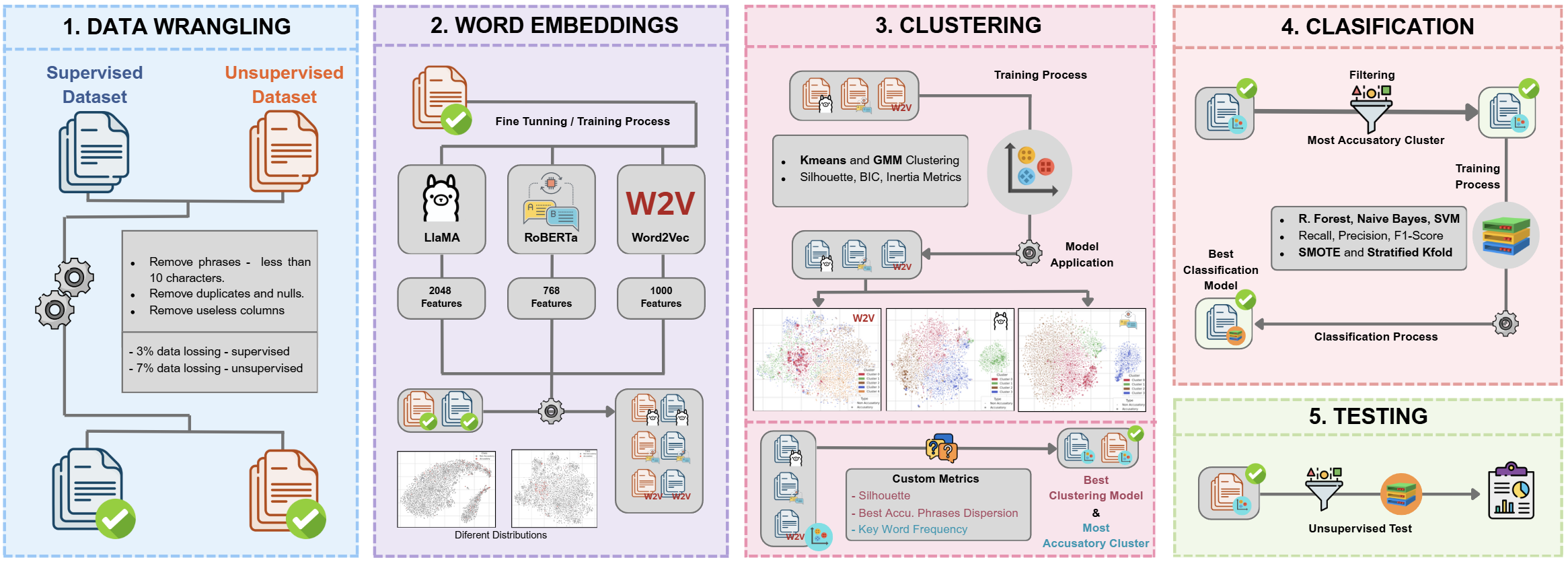}
    \caption{Block diagram of the proposed approach.}
    \label{fig:example}
\end{figure*}
This section describes the datasets and their sources, the preprocessing steps applied to the raw data, the models and algorithms employed, and the relevant implementation details. An overview of the complete methodological workflow is shown in Fig.~\ref{fig:example}, which summarizes the proposed end-to-end pipeline. The methodology is organized into the following stages: data wrangling, embedding generation, clustering, cluster identification, classification, and model testing. The proposed workflow is tailored to the problem of detecting accusatory content in public procurement data. Although prior studies have addressed corruption detection, no standardized pipeline exists for handling this type of information. Methodological choices at each stage were informed by related work and designed to support efficient semi-supervised learning under large-scale unlabeled data and highly imbalanced labeled datasets.

\subsection{Data Wrangling}
The datasets used in this study were obtained through the Kapak framework, which automates the collection of public procurement data from Ecuador’s SOCE. The extracted information is stored in the Kapak database and organized into two subsets: a labeled dataset with ground-truth annotations and a substantially larger unlabeled dataset. {In particular, the supervised dataset was manually reviewed by three annotators from the Kapak project team using a task-specific binary annotation guideline. A question or comment was labeled as accusatory only when it explicitly denoted corruption, bid manipulation, directed procurement, illegality, prohibited conditions, or intentional procedural irregularities within a public procurement process. Conversely, generic complaints, administrative questions, expressions of dissatisfaction without an explicit accusation, or ambiguous statements were labeled as non-accusatory.} {The final label for each sample was assigned using a majority-vote criterion among the three annotators. Therefore, a sentence was labeled as accusatory when at least two annotators agreed on the accusatory label. This annotation procedure resulted in a final supervised dataset of 5,005 questions, of which 143 were labeled as accusatory. To assess the consistency of the annotation process, inter-annotator agreement (IAA)} \cite{artstein-poesio-2008-survey} {was measured using Fleiss' Kappa, which is suitable for nominal labels assigned by more than two annotators. The obtained value was $\kappa = 0.72$, indicating a substantial level of agreement among the three annotators. Fleiss' Kappa was computed as $\kappa = \frac{\bar{P} - \bar{P}_{e}}{1 - \bar{P}_{e}}$, where $\bar{P}$ denotes the observed agreement and $\bar{P}_{e}$ denotes the agreement expected by chance.} {For this study, accusatory language was operationally defined as a question or comment that explicitly attributes or strongly implies irregular conduct within a public procurement process. This operational definition was used to distinguish accusatory language from general dissatisfaction, complaints, administrative questions, or strong negative opinions. The inclusion and exclusion criteria applied during the annotation process are summarized in } Table~\ref{tab:accusatory_definition}.

\begin{table}[!t]
\centering
\small
\caption{{Operational Criteria for Accusatory Language Annotation}}
\label{tab:accusatory_definition}
\begin{tabular}{p{2.8cm}p{5.1cm}}
\hline
\textbf{ {Category}} & \textbf{{Annotation Criterion}} \\
\hline
\textbf{ {Accusatory}} & 
{A question or comment was labeled as accusatory when it explicitly attributed or strongly implied irregular conduct within a public procurement process, including corruption, bid manipulation, directed procurement, illegality, prohibited conditions, intentional procedural bias, or unfair influence over the process.} \\
\hline
\textbf{{Non-accusatory}} & 
{A question or comment was labeled as non-accusatory when it expressed general dissatisfaction, disagreement, frustration, urgency, criticism of technical specifications, administrative uncertainty, requests for clarification, or strong negative opinions without an explicit allegation or clear implication of corruption or procedural manipulation.} \\
\hline
\end{tabular}
\end{table}

These datasets support the supervised and unsupervised learning stages of the proposed pipeline and were subjected to an initial preprocessing stage prior to analysis.
Overall, the datasets were relatively clean\footnote{In this context, ``clean'' refers to the absence of empty records, entries composed solely of special characters, or encoding-related corruption (e.g., malformed diacritics).}, requiring only minimal preprocessing to ensure consistency and relevance.
\begin{table}[!t]
\caption{Summary of Dataset Characteristics}
\label{tab:gen_dataset_data}
\footnotesize
\begin{tabular*}{20.25pc}{@{}lcccr@{}}
\hline\\[-1.5pc]\hline
\textbf{Type} &
\textbf{Rows} &
\textbf{Col. Num.} &
\textbf{Min Ch.} &
\textbf{Max Ch.} \\
\hline
Unsupervised & 100,000 & 3 & 1 & 990 \\
Supervised   & 5,005 & 4 & 6 & 990 \\
\hline\\[-1.5pc]\hline
\end{tabular*}
\end{table}

Beyond the basic metrics reported in Table~\ref{tab:gen_dataset_data}, the datasets exhibit several relevant characteristics. Question lengths range from 1 to 990 characters, reflecting high variability in textual content. The supervised dataset is strongly imbalanced, with accusatory questions accounting for approximately 3\% of the samples (143 out of 5,005). Additionally, some metadata and transactional fields were excluded, as they were not informative for the NLP tasks considered.

To prepare the data for modeling, several preprocessing and transformation steps were applied. Questions shorter than 10 characters were removed due to their limited informational value, and entries composed solely of stopwords, articles, or short words were discarded. Duplicate questions were eliminated to prevent training bias, and only relevant columns were retained; for the supervised dataset, this included the question text and its corresponding accusatory label. These preprocessing steps resulted in only minor data reduction, with no loss of accusatory instances. By removing low-value and noisy content, the overall quality of the training data was improved. A summary of the data retention and filtering process is presented in Table~\ref{tab:datasets_specs}.

\begin{table}[t!]
\caption{Summary of Data Cleaning Results}
\label{tab:datasets_specs}
\footnotesize
\begin{tabular*}{20.25pc}{@{}lccc@{}}
\hline\\[-1.5pc]\hline
\textbf{Dataset} &
\makecell{\textbf{Initial Rows}} &
\makecell{\textbf{Cleaned Rows}} &
\makecell{\textbf{Data Loss}} \\
\hline
Unsupervised & 100,000 & 92,579 & 7.42\% \\
Supervised   & 5,005 & 4,841 & 3.28\% \\
\hline\\[-1.5pc]\hline
\end{tabular*}
\end{table}
\subsection{Embedding Generation}
Textual data must be transformed into numerical representations to enable computational analysis. Text embeddings encode sentences or documents as dense vectors in an $n$-dimensional space, allowing machine learning models to operate on linguistic content \cite{birunda2021review}. In this study, we evaluate three embedding strategies: two Transformer-based large language models (i.e., LLaMA 3.2 (1B parameters) \cite{meta2024llama32} and RoBERTa \cite{liu2019roberta}) and a classical Word2Vec architecture. These embedding techniques are widely adopted in NLP tasks and have demonstrated strong performance in text classification problems based on both semantic and structural features \cite{txt_class_w_vec_wang_yan,zhao2025advancingsinglemultitasktext,txt_class_w2v_xiao_wang,sv_w2v_txt_class_lilllebert_zhu,Santos2025}. {In this study, the fine-tuned RoBERTa and LLaMA models were considered as adapted Transformer-based baselines for evaluating whether domain-specific fine-tuning could improve sentence-level embedding quality for downstream clustering. Therefore, these models were not intended to represent fully optimized large-scale LLM embedding systems, but rather controlled Transformer-based alternatives within the proposed end-to-end pipeline. This distinction is important because obtaining highly effective sentence embeddings from Transformer-based models often requires additional adaptation strategies, such as larger domain-specific corpora, contrastive learning objectives, or architectures explicitly designed for sentence representation learning.} Before being used in downstream tasks, text embeddings must be extracted in a consistent manner across models. For this purpose, a unified embedding extraction strategy was adopted for all language models.

\textbf{Embedding Extraction Strategy:} Mean pooling was applied to generate fixed-length sentence representations by averaging token-level embeddings while excluding padding tokens. This approach provides a simple and effective aggregation mechanism and is widely used in downstream NLP tasks \cite{abs-1908-10084}.
\begin{figure}[t!]
    \centering
    \includegraphics[width=\columnwidth]{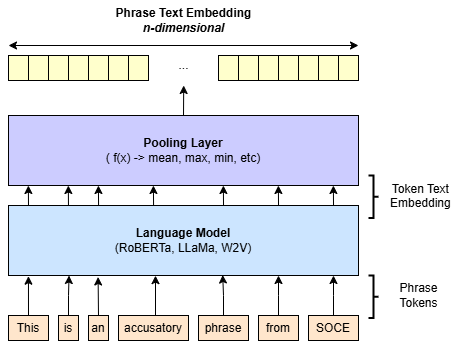}
    \caption{General overview of the mean pooling operation applied across all models.}
    \label{fig:meanpooling_diagram}
\end{figure}
As illustrated in Fig.~\ref{fig:meanpooling_diagram}, each input phrase is tokenized and encoded by the corresponding language model (RoBERTa, LLaMA, or Word2Vec), producing token-level embeddings that are subsequently aggregated via mean pooling. The dimensionality of the resulting sentence embeddings depends on the hidden size of each model and is accounted for in subsequent clustering and classification stages.

\subsection{Clustering}

After generating embeddings for both the supervised and unsupervised datasets (Table~\ref{tab:datasets_specs}), the next step is to uncover latent structures in the textual data. The objective of clustering is to identify regions in the unlabeled dataset that are likely to contain accusatory content, thereby guiding subsequent classification. Given the strong class imbalance in the labeled data, a similar imbalance is assumed in the unlabeled corpus, motivating a targeted clustering strategy. Accordingly, two widely used clustering algorithms were evaluated: k-means, a hard-assignment method \cite{lloyd1982least}, and Gaussian Mixture Models (GMMs), which provide soft assignments \cite{scikit-learn-gmm}.

To determine the most suitable clustering configuration, a structured model selection strategy was employed.

\textbf{Model Selection Strategy:}  
Two complementary criteria were used to select both the clustering method and the number of clusters:

\begin{itemize}
    \item \textit{Elbow Method:} Clustering performance was evaluated as a function of the number of clusters $k$ using metrics such as inertia and silhouette score, identifying diminishing performance gains beyond an optimal $k$ \cite{Bholowalia2014EBKMeans}.
    \item \textit{Supervised Evaluation:} The labeled dataset was used to assess whether accusatory phrases were concentrated within specific clusters, considering factors such as cluster density, purity, and the distribution of positive samples.
\end{itemize}

Clustering performance was quantified using the following evaluation metrics:

\begin{itemize}
    \item \textit{Silhouette Score:} Measures intra-cluster similarity relative to inter-cluster separation \cite{rousseeuw1987silhouettes}.
    \item \textit{MNAPK:} Maximum number of accusatory phrases contained in a single cluster.
    \item \textit{TNCK:} Total number of samples assigned to cluster $K$.
    \item \textit{Relative accusatory frequency (cluster-level):} Proportion of accusatory samples within cluster $K$.
    \item \textit{Relative accusatory frequency (dataset-level):} Proportion of all accusatory samples captured by cluster $K$.
\end{itemize}

\subsection{Cluster Identification}\label{sec:cluster_identification}
After clustering, it is necessary to identify which cluster is most likely to contain accusatory content in order to prioritize data segments for further analysis. To this end, we adopt a keyword-based identification strategy applied to both the labeled and unlabeled datasets. Keyword-based methods have shown strong performance in topic and intent identification, particularly in short and noisy text scenarios and under limited supervision or class imbalance \cite{soc_med_clust_comito_2019}.

The proposed strategy assumes that accusatory questions exhibit specific lexical cues. Based on domain expertise and exploratory analysis, the following set of high-impact keywords was selected:

\begin{itemize}
    \item \textit{Direccionado} (Directed procurement) 
    \item \textit{Imposible} (Impossible)
    \item \textit{Corrupción} (Corruption)
    \item \textit{Ilegal} (Illegal)
    \item \textit{Prohibido} (Prohibited)
\end{itemize}

\begin{figure}[!t]
    \centering
    \includegraphics[width=\columnwidth]{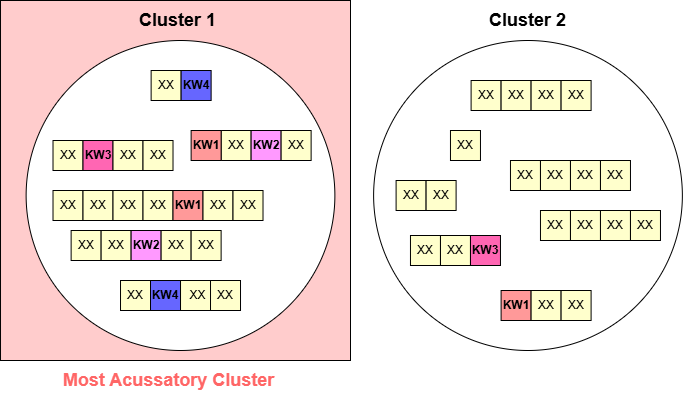}
    \caption{Strategy for identifying the most accusatory cluster.}
    \label{fig:clus_desc_est}
\end{figure}

The cluster with the highest cumulative frequency of at least four of these five keywords is identified as the most accusatory cluster. This procedure is performed independently on both datasets to ensure robustness and scalability. Fig.~\ref{fig:clus_desc_est} summarizes the identification process.

\subsection{Classification}

This stage evaluates the effectiveness of different classifiers for detecting accusatory content. Since labeled data is required, classification is performed exclusively on the supervised dataset. The classification pipeline consists of the following steps:

\begin{enumerate}
    \item \textbf{Cluster Filtering:} Records belonging to the cluster identified as the most accusatory are extracted, concentrating positive samples and reducing noise.
    \item \textbf{Model Training and Evaluation:} Classification models are trained and evaluated using the filtered dataset, mitigating class imbalance and enabling a scalable semi-supervised setup.
\end{enumerate}

Despite filtering, class imbalance may persist. To reduce its impact and limit classification bias, the following techniques are applied:

\begin{itemize}
    \item \textbf{SMOTE:} Synthetic minority samples are generated through interpolation, improving class balance without data duplication \cite{chawla2002smote}.
    \item \textbf{Stratified K-Fold Cross-Validation:} The dataset is partitioned into $k$ folds while preserving class proportions, ensuring robust and unbiased evaluation \cite{scikit-learn_stratifiedkfold}.
\end{itemize}

For the binary classification task, three classifiers representing different methodological paradigms were selected based on prior success in text classification: Random Forest \cite{breiman2001random}, Gaussian Naive Bayes (GNB) \cite{n_bayes_2018_ieee}, and Support Vector Machine (SVM) \cite{ZHANG2008879}.
\begin{figure}[!t]
    \centering
    \includegraphics[width=\columnwidth]{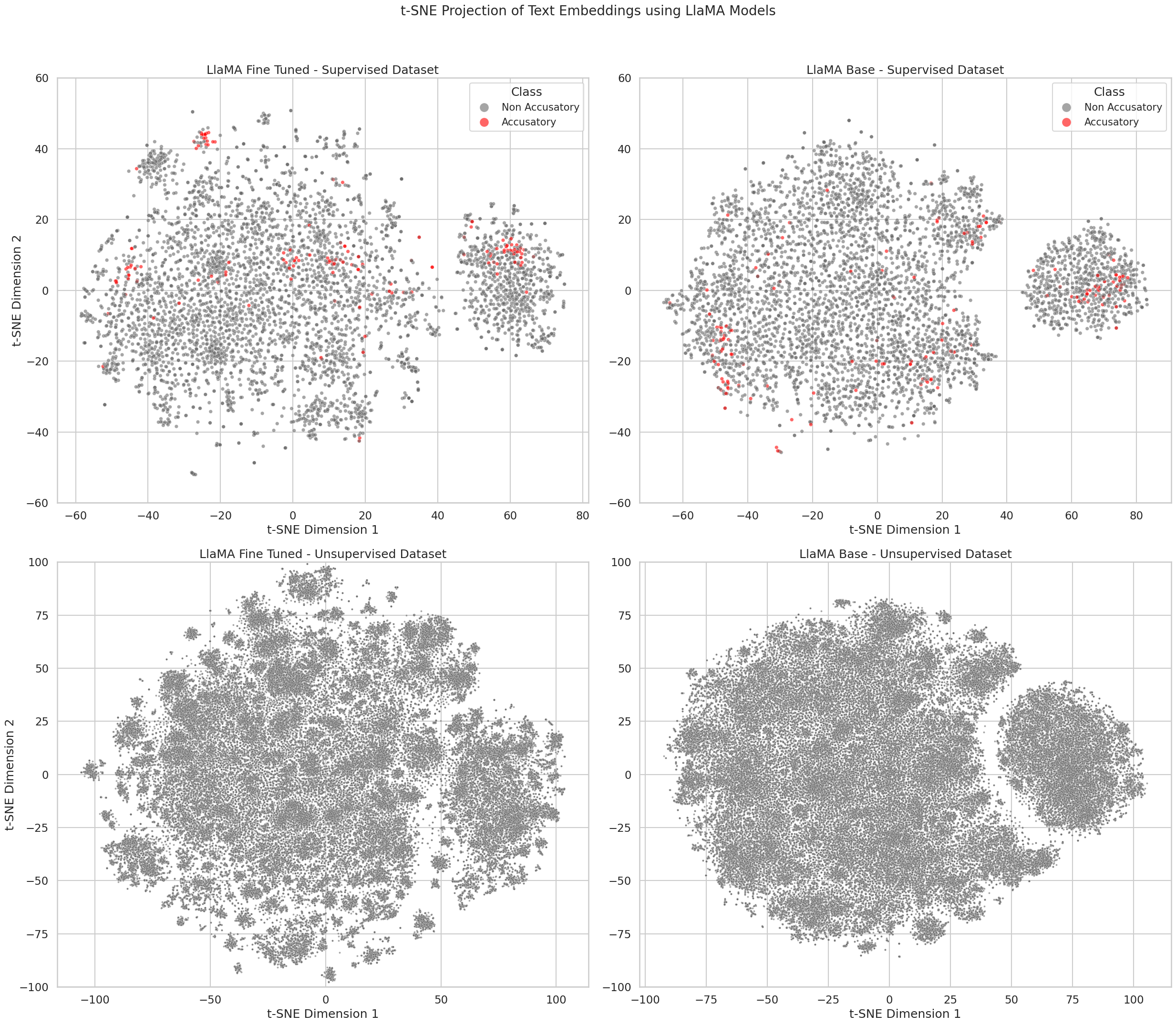}
    \caption{t-SNE projection of LLaMA embeddings for accusatory phrases.}
    \label{fig:tsne_llama}
\end{figure}
\subsection{Model Testing}

The final stage of the proposed pipeline applies the trained classifier to the unlabeled dataset in order to evaluate its practical usefulness in scenarios where labeled data is scarce or unavailable. Model testing follows a structured workflow to ensure consistency and reproducibility:

\begin{enumerate}
    \item \textbf{Cluster Identification:} Identify the most accusatory cluster within the unlabeled dataset using the keyword-based method described in Section~\ref{sec:cluster_identification}.
    \item \textbf{Data Filtering:} Extract records belonging to the selected cluster, which is expected to concentrate a higher proportion of accusatory content.
    \item \textbf{Classification:} Apply the pre-trained classifier obtained during the supervised phase to generate predictions on the filtered data.
    \item \textbf{Information Extraction:} Collect predicted positive samples to support validation, reporting, and feedback processes within the Kapak framework.
\end{enumerate}

This stage enables the detection of potentially irregular or accusatory comments without manual labeling, supporting a semi-automated auditing workflow and providing early warning signals for procurement oversight. Additional details on training, testing, and implementation are available in the project repository at \href{https://github.com/Espejin/USFQ-MSDS-Accusatory-Comments-Classification}{\textit{$<$Kapak NLP Pipeline$>$}}.
\begin{figure}[!t]
    \centering
    \includegraphics[width=\columnwidth]{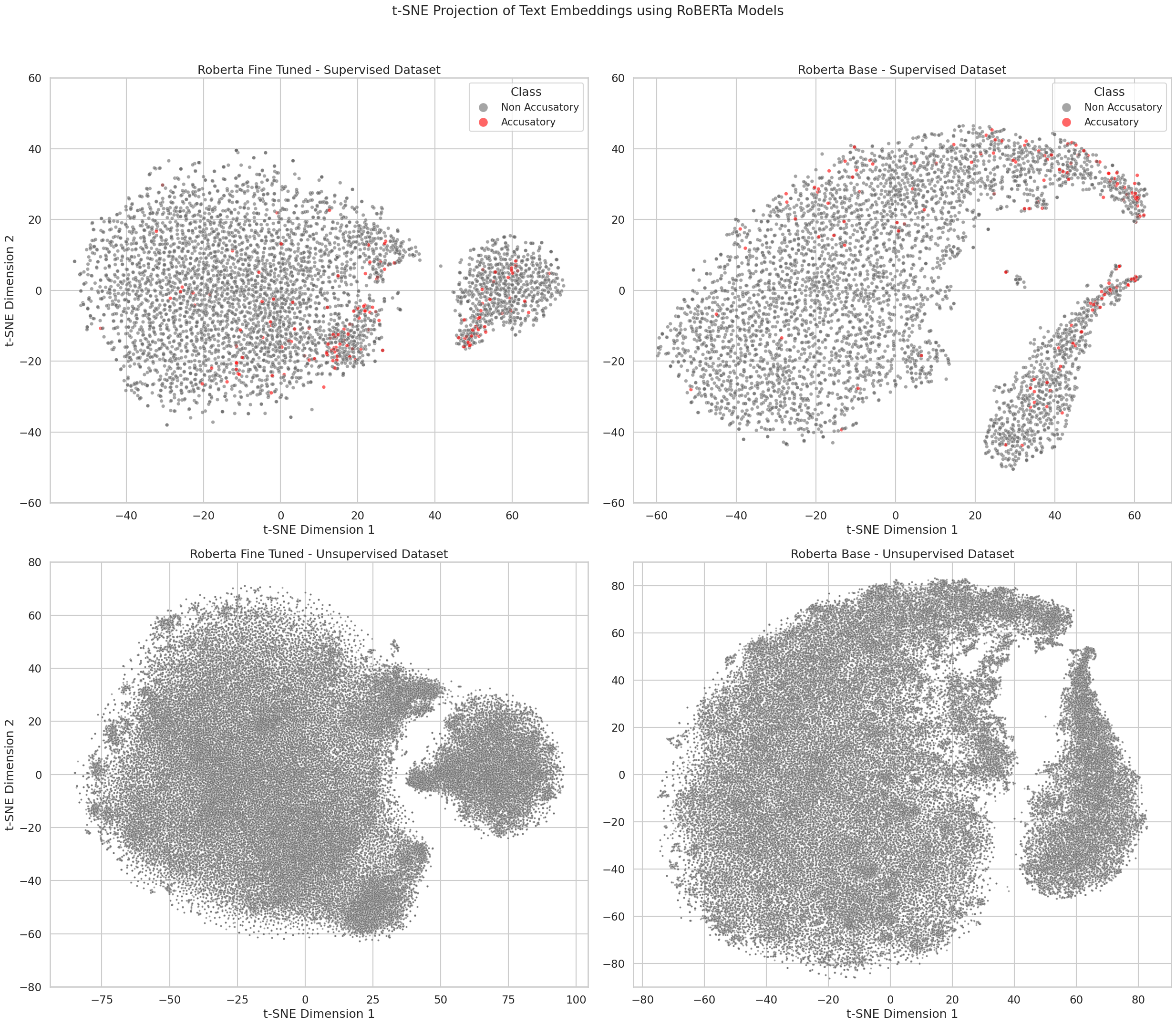}
    \caption{t-SNE projection of RoBERTa embeddings for accusatory phrases.}
    \label{fig:tsne_roberta}
\end{figure}
\section{Results and Discussion}
\label{sec:results}
This section analyzes the results obtained from the proposed hybrid methodology, focusing on the performance of its main components: embedding generation, clustering of unlabeled data, identification of accusatory clusters, supervised classification, and large-scale testing. The discussion emphasizes the impact of different embedding strategies, clustering configurations, keyword-based cluster identification, and classification performance under class imbalance, as well as the robustness of the framework in unsupervised settings.

\subsection{Word Embeddings}

Sentence embeddings were generated for both supervised and unsupervised datasets using three techniques: LLaMA, RoBERTa, and Word2Vec. These models differ in architecture, dimensionality, and semantic representation capacity. To examine the latent structure of the embeddings and assess semantic separability, dimensionality reduction was performed using t-SNE, enabling a qualitative analysis of clustering tendencies, particularly for accusatory phrases.

\begin{table*}[t]
\caption{Clustering Evaluation Results -- Most Promising Models}
\label{tab:clustering_results}
\footnotesize
\setlength{\tabcolsep}{7pt}
\centering
\begin{tabular*}{\textwidth}{@{}l l c c c c c c@{}}
\hline
\makecell{\textbf{Clustering} \\ \textbf{Method}} & 
\makecell{\textbf{NLP} \\ \textbf{Method}} & 
\makecell{\textbf{Cluster} \\ \textbf{Number}} & 
\makecell{\textbf{Silhouette} \\ \textbf{Score}} & 
\textbf{MNAPK} & 
\textbf{TNCK} & 
\makecell{\textbf{Rel. Freq.} \\ \textbf{Acu. Phrases} \\ \textbf{Over TNCK}} & 
\makecell{\textbf{Rel. Freq.} \\ \textbf{Acu. Phrases} \\ \textbf{Over TNAP}} \\
\hline
 & LLaMA 3.2-1B Fine Tune & 6 & -0.023 & 47 & 1498 & 3.14\% & 32.87\% \\
 & LLaMA 3.2-1B Base      & 6 & 0.108  & 44 & 782  & 5.63\% & 30.77\% \\
GMM & RoBERTa Fine Tune   & 5 & 0.079  & 64 & 1149 & 5.57\% & 44.76\% \\
 & RoBERTa Base          & 5 & 0.112  & 50 & 866  & 5.77\% & 34.97\% \\
 & \textbf{Word2Vec}     & 5 & 0.273  & 122 & 962 & 12.68\% & 85.31\% \\
\hline
 & LLaMA 3.2-1B Fine Tune & 4 & 0.075 & 45 & 1150 & 3.91\% & 31.47\% \\
 & LLaMA 3.2-1B Base      & 4 & 0.149 & 45 & 1106 & 4.07\% & 31.47\% \\
KMeans & RoBERTa Fine Tune & 5 & 0.105 & 73 & 1391 & 5.25\% & 51.05\% \\
 & RoBERTa Base          & 4 & 0.198 & 58 & 440 & 13.18\% & 40.56\% \\
 & \textbf{Word2Vec}     & 5 & 0.059 & 129 & 1276 & 10.11\% & 90.21\% \\
\hline
\end{tabular*}

\vspace{2pt}
\footnotesize
\textbf{TNAP}: Total number of accusatory phrases in the labeled dataset. \;
\textbf{TNCK}: Total number of comments in the cluster. \;
\textbf{MNAPK}: Number of accusatory phrases in the cluster. \;
\textbf{Rel. Freq. Over TNCK}: Proportion of accusatory phrases within the cluster. \;
\textbf{Rel. Freq. Over TNAP}: Proportion of all accusatory phrases captured by the cluster.
\end{table*}

\subsubsection{LLaMA}
The LLaMA 3.2-1B model produced 2048-dimensional embeddings per phrase. As observed in Fig.~\ref{fig:tsne_llama}, LLaMA embeddings exhibit a scattered distribution across the projection space. While some localized groupings emerge, the majority of accusatory phrases remain widely dispersed. The fine-tuned version of the model slightly alters the spatial layout, suggesting some degree of semantic adaptation to the domain-specific dataset. However, the presence of compact accusatory clusters is limited.

\subsubsection{RoBERTa}
RoBERTa generated 768-dimensional embeddings per sentence. As shown in Fig.~\ref{fig:tsne_roberta}, the embeddings produced by RoBERTa form several clusters with slightly more compactness than those observed with LLaMA. The impact of fine-tuning is evident, as the spatial configuration becomes more structured and exhibits better separation. This is likely a result of adapting the model, originally pre-trained on English text, to the Spanish domain of the Kapak corpus.

\subsubsection{Word2Vec}
Using Word2Vec, we obtained 1000-dimensional embeddings for each phrase. Unlike the transformer-based methods, Word2Vec was trained directly on the Kapak corpus, incorporating a custom tokenizer to handle domain-specific characteristics such as misspellings and colloquialisms.
\begin{figure}[!t]
    \centering
    \includegraphics[width=\columnwidth]{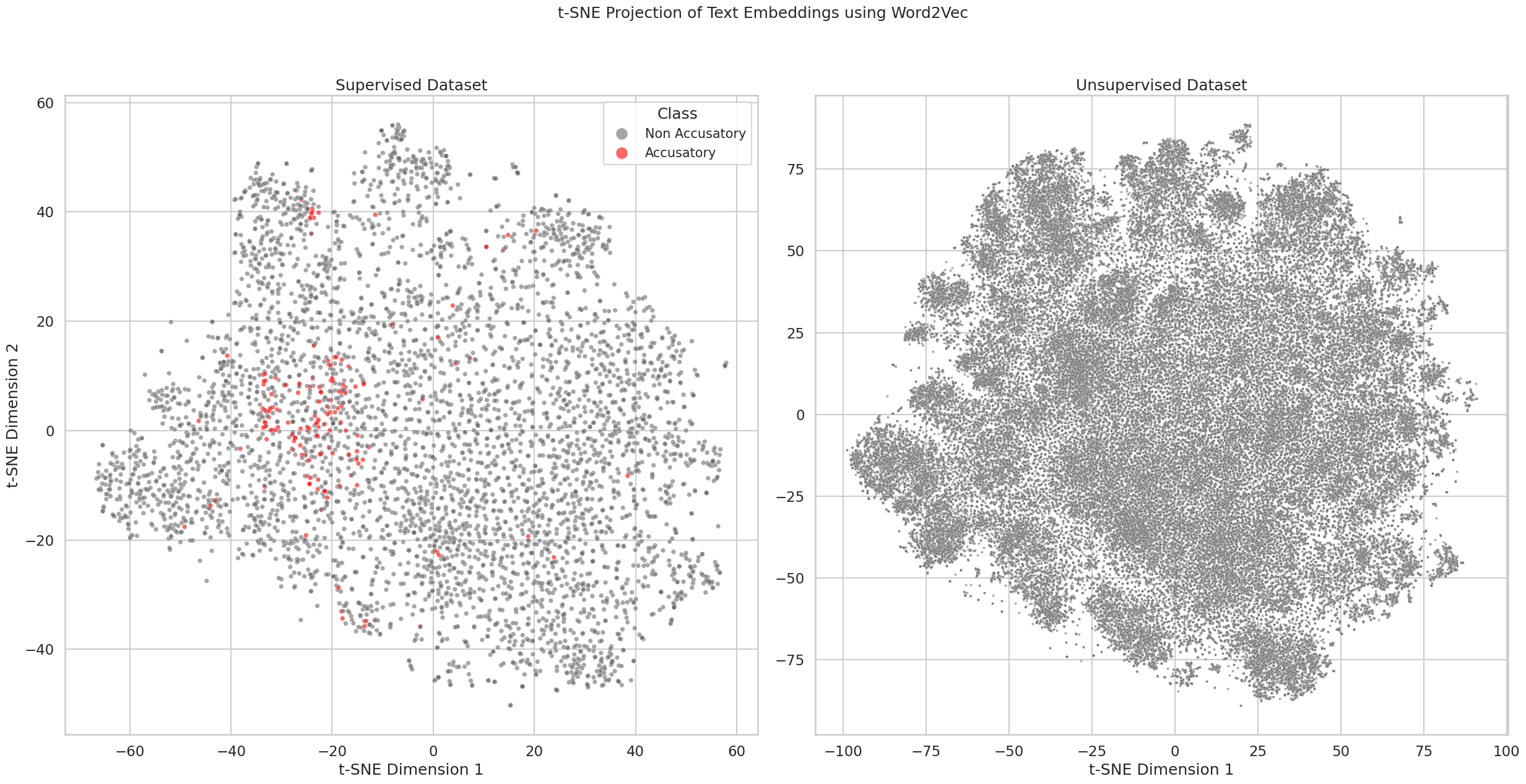}
    \caption{t-SNE projection of Word2Vec embeddings for accusatory phrases.}
    \label{fig:tsne_w2v}
\end{figure}
Fig.~\ref{fig:tsne_w2v} reveals a denser concentration of embeddings in specific regions of the space. Unlike LLaMA and RoBERTa, which exhibit broader semantic dispersion, Word2Vec shows a localized accumulation of accusatory phrases, indicating its potential utility in downstream clustering and classification tasks.
\noindent These results reveal meaningful insights:
\begin{itemize}
    \item Transformer-based embeddings (LLaMA and RoBERTa) are more dispersed, reflecting their richer contextual representations.
    \item RoBERTa shows considerable spatial reorganization post fine-tuning, likely due to its monolingual pretraining in English.
    \item Word2Vec, although less expressive, generates compact regions that facilitate cluster formation and labeling of accusatory content.
\end{itemize}
These structural differences in embedding space have direct implications for clustering strategy, influencing the segmentation and classification of accusatory phrases in subsequent stages of the pipeline.
\subsection{Clustering}
Clustering plays a pivotal role in this study, as it allows the grouping of similar questions based on their semantic representations, without relying on class labels. This process is especially valuable in a semi-supervised context, as it helps identify dense regions of accusatory content that can be prioritized for further classification. Here, we evaluate different clustering strategies across various embedding methods to determine the most effective approach for isolating accusatory phrases.

\textbf{Model Selection Strategy.} As outlined in previous sections, a hybrid approach was employed to determine the optimal clustering model. This strategy combined unsupervised validation metrics with domain-specific criteria, particularly focusing on the enrichment of accusatory phrases within identified clusters. Table~\ref{tab:clustering_results} presents a detailed summary of the top-performing models according to metrics such as Silhouette Score, Maximum Number of Accusatory Phrases in a Cluster (MNAPK), and their relative frequencies. Although Silhouette scores were generally low, this is consistent with the behavior of high-dimensional word embeddings, which tend to produce overlapping and complex data structures. Literature in the field has noted similar behavior when applying clustering to embeddings \cite{saha2023influence, wu2021comprehensive}.

{Although the Silhouette scores obtained in this study are below the values typically expected for well-separated low-dimensional datasets, this behavior is consistent with the characteristics of high-dimensional text embedding spaces. In such spaces, semantic representations often exhibit overlapping structures, anisotropy, and complex non-spherical distributions. Consequently, distance-based compactness and separation metrics may underestimate the practical usefulness of a clustering solution.} {The low performance observed in metrics such as the Silhouette score is also attributed to a phenomenon known as distance concentration. This principle establishes that, as the dimensionality of a vector space increases, the distances between any given points tend to converge to a nearly identical value} \cite{peng2025interpretingcursedimensionalitydistance}. {Consequently, algorithms such as K-Means may struggle to assign distinct samples to specific clusters, as the distance from a data point to multiple competing centroids becomes practically indistinguishable. Similarly, probabilistic models such as GMM may fail to decisively determine cluster membership when posterior probabilities become less differentiated across clusters.}  {For this reason, the Silhouette score was not used as the sole criterion for selecting the best clustering configuration. Instead, complementary internal metrics were considered, including Inertia for K-Means and the Bayesian Information Criterion (BIC) for GMM. In addition, task-oriented metrics were incorporated, such as the maximum number of accusatory phrases in a cluster (MNAPK), the relative frequency of accusatory phrases within the selected cluster, and the proportion of total accusatory phrases captured by that cluster. This evaluation strategy is aligned with the objective of the clustering stage, which is not to generate a perfectly separated taxonomy of all comments, but to identify a semantically meaningful region enriched with accusatory content for subsequent classification. Therefore, the low Silhouette scores do not invalidate the utility of the clustering stage; rather, they reflect the limitations of distance-based validation metrics in high-dimensional embedding spaces.}


\subsubsection{Clustering over LLaMA embeddings}

Clustering based on LLaMA embeddings exhibited high dispersion across the embedding space. Although one cluster captured a non-negligible number of accusatory phrases, their relative frequency within the cluster remained low, indicating limited precision in isolating accusatory content.
\begin{figure}[!t]
    \centering
    \includegraphics[width=\columnwidth]{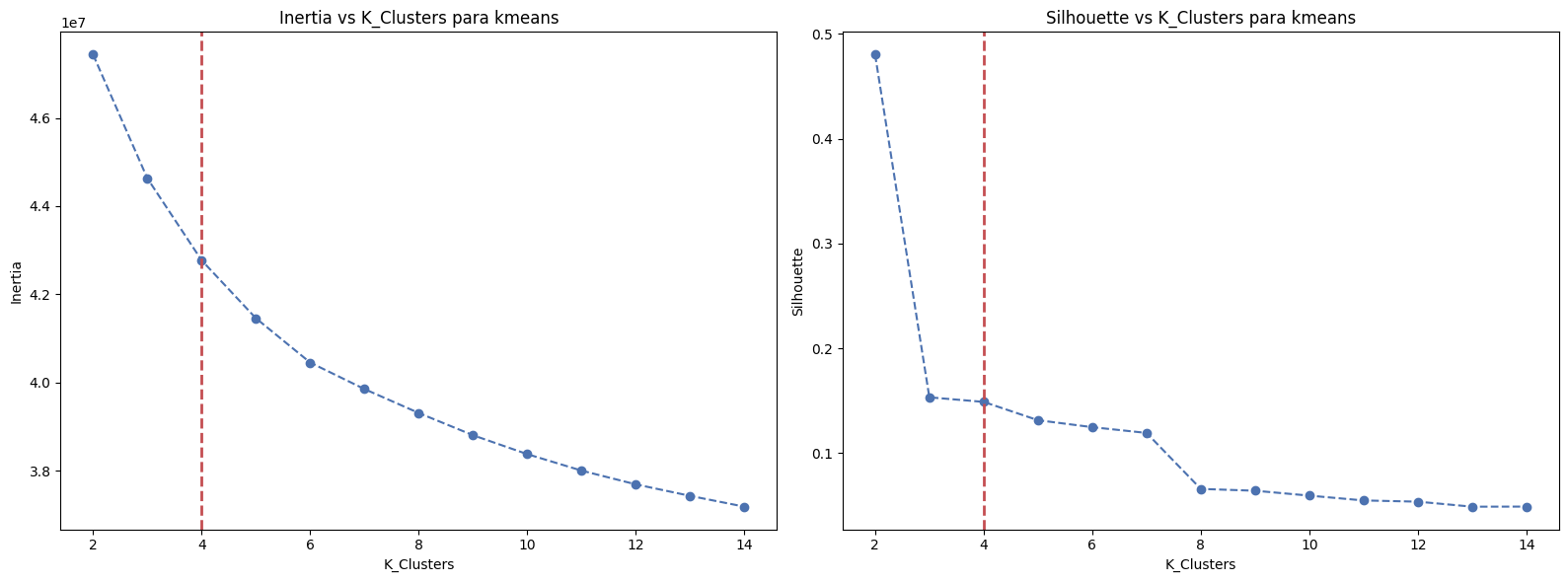}
    \caption{Inertia and Silhouette metrics for KMeans using LLaMA 3.2-1B Base embeddings.}
    \label{fig:in_sil_km_llm}
\end{figure}
\begin{figure}[!t]
    \centering
    \includegraphics[width=\columnwidth]{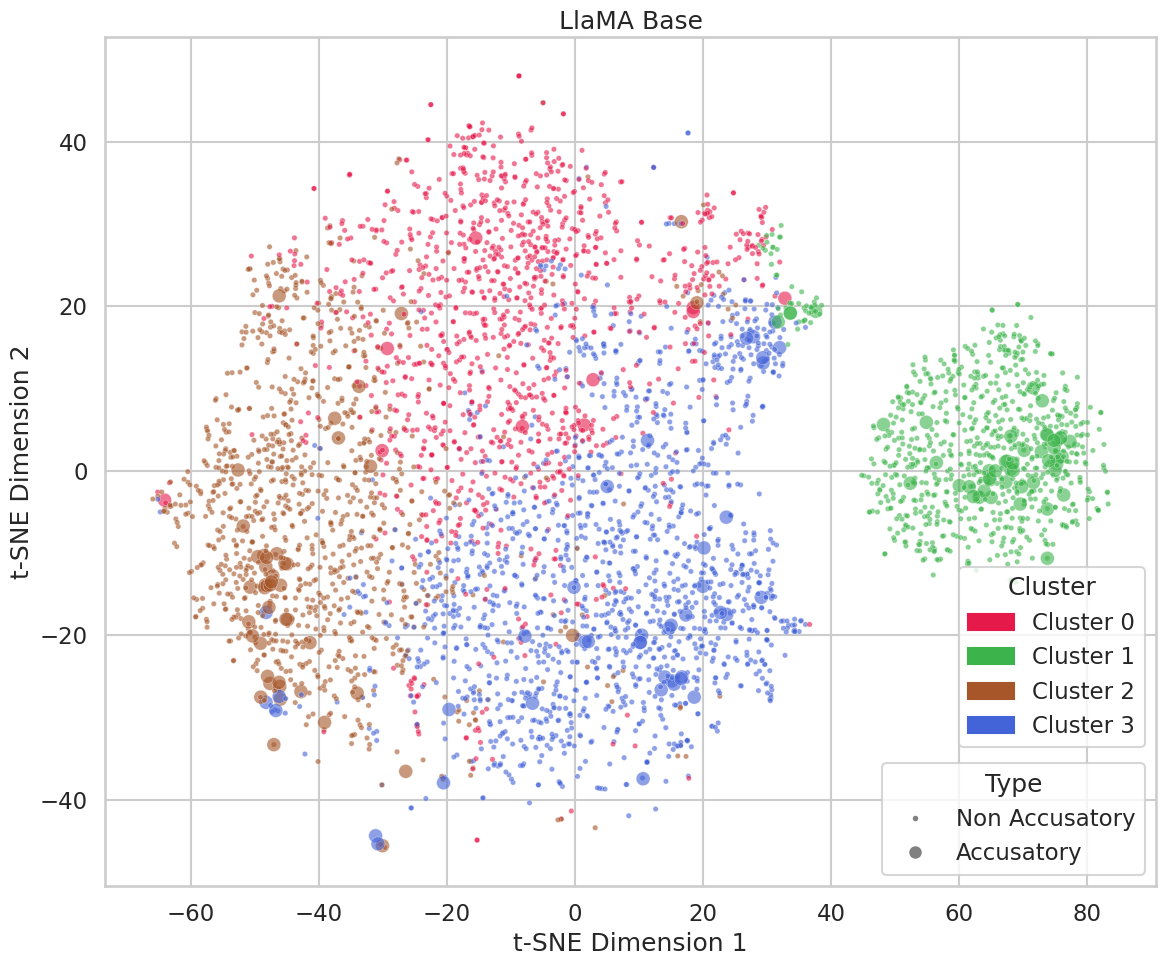}
    \caption{t-SNE projection of the supervised dataset clustered with KMeans using LLaMA 3.2-1B Base embeddings.}
    \label{fig:llm_bse_clst}
\end{figure}
As shown in Fig.~\ref{fig:in_sil_km_llm}, increasing the number of clusters did not lead to meaningful improvements in clustering quality, as reflected by consistently low Silhouette scores. The t-SNE visualization in Fig.~\ref{fig:llm_bse_clst} further illustrates that, although cluster 2 contained the highest number of accusatory phrases, these instances were widely dispersed, limiting the effectiveness of LLaMA-based clustering for downstream classification tasks.

\subsubsection{Clustering over RoBERTa embeddings}

RoBERTa-based embeddings showed improved clustering performance compared to LLaMA, with several configurations concentrating more than 50\% of the accusatory phrases within a single cluster, indicating stronger semantic cohesion.
\begin{figure}[!t]
    \centering
    \includegraphics[width=\columnwidth]{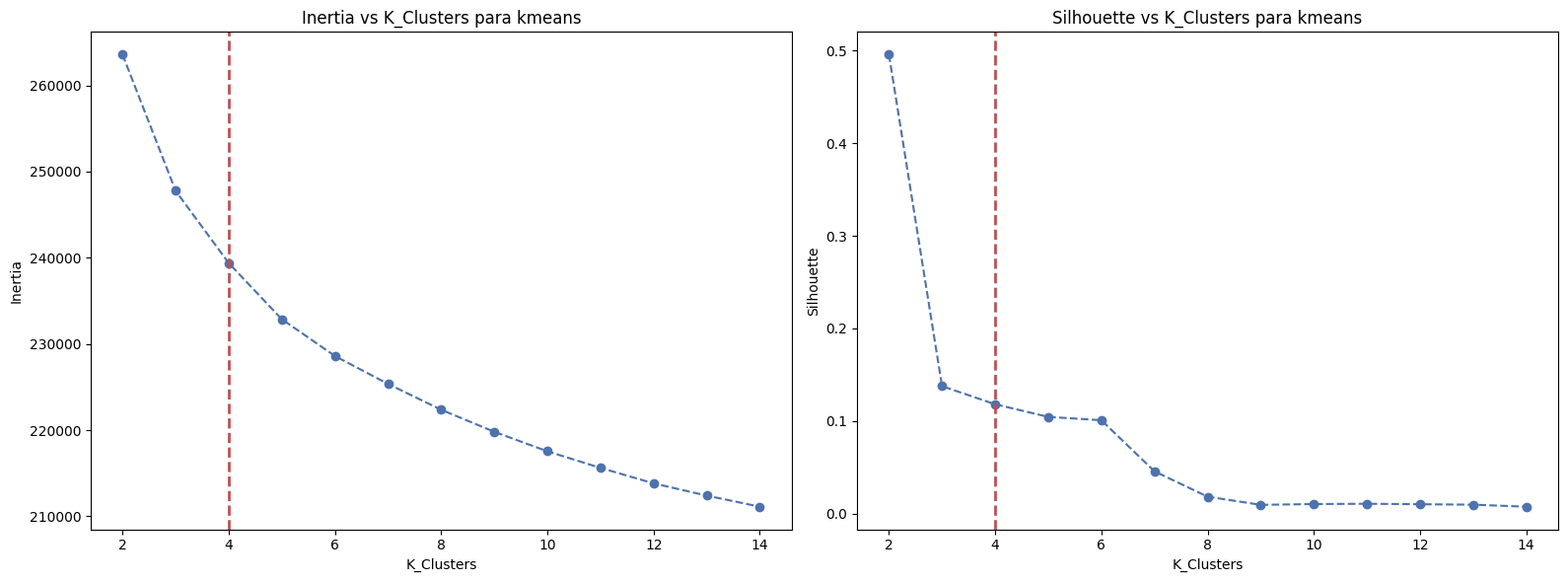}
    \caption{Inertia and Silhouette metrics for KMeans using RoBERTa fine-tuned embeddings.}
    \label{fig:in_sil_km_rob}
\end{figure}
As observed in Fig.~\ref{fig:in_sil_km_rob}, the highest Silhouette score was obtained at $k=3$, beyond which semantic cohesion degraded despite continued reductions in inertia.
\begin{figure}[!t]
    \centering
    \includegraphics[width=\columnwidth]{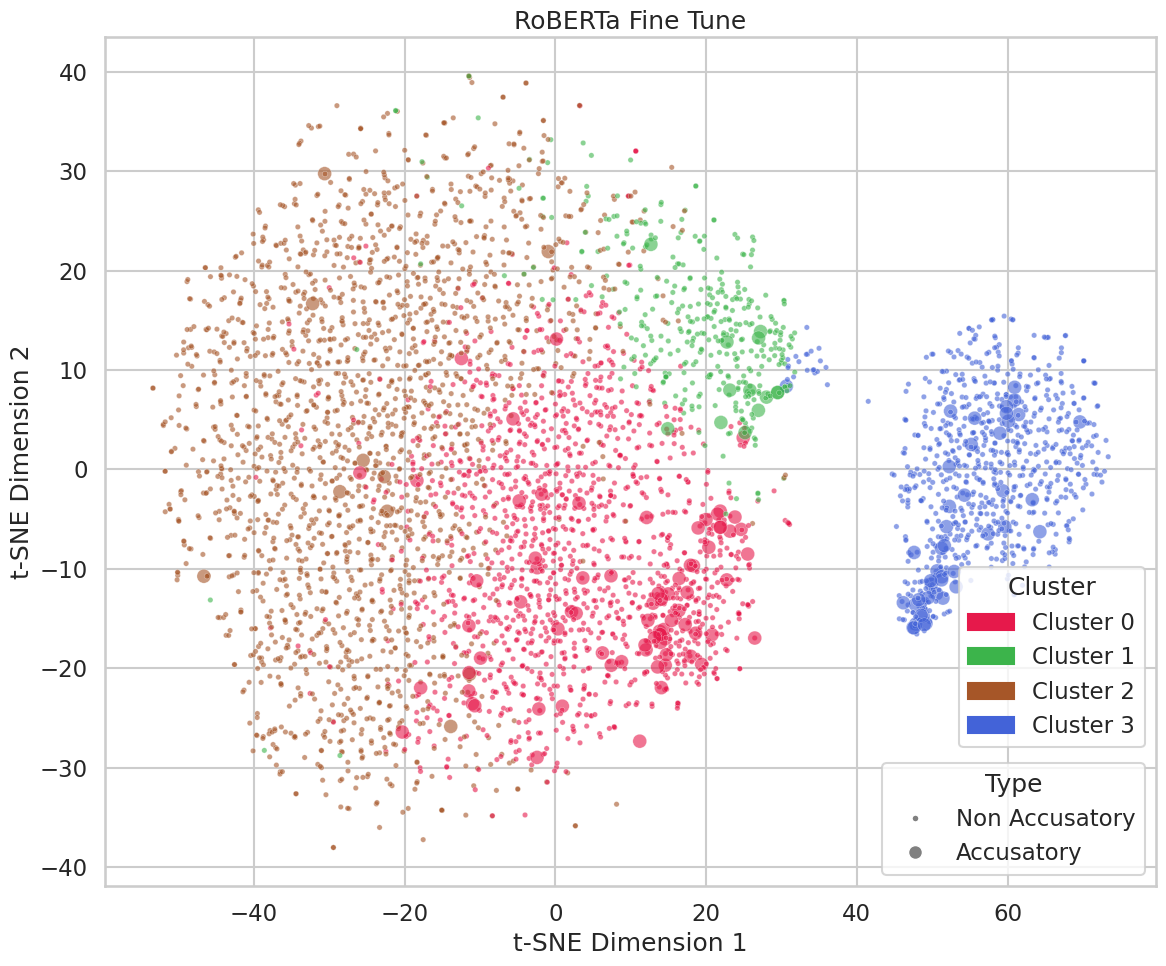}
    \caption{t-SNE projection of the supervised dataset clustered with KMeans using RoBERTa fine-tuned embeddings.}
    \label{fig:rob_fn_clst}
\end{figure}
The t-SNE visualization in Fig.~\ref{fig:rob_fn_clst} shows that cluster 0 captured most accusatory instances with significantly less dispersion than LLaMA, suggesting that fine-tuning enhanced RoBERTa’s suitability for accusatory content detection.

\subsubsection{Clustering over Word2Vec embeddings}
Among all evaluated methods, the combination of Word2Vec embeddings and GMM achieved the best clustering performance, yielding the highest Silhouette score and the greatest concentration of accusatory phrases within a single cluster.
\begin{figure}[!t]
    \centering
    \includegraphics[width=\columnwidth]{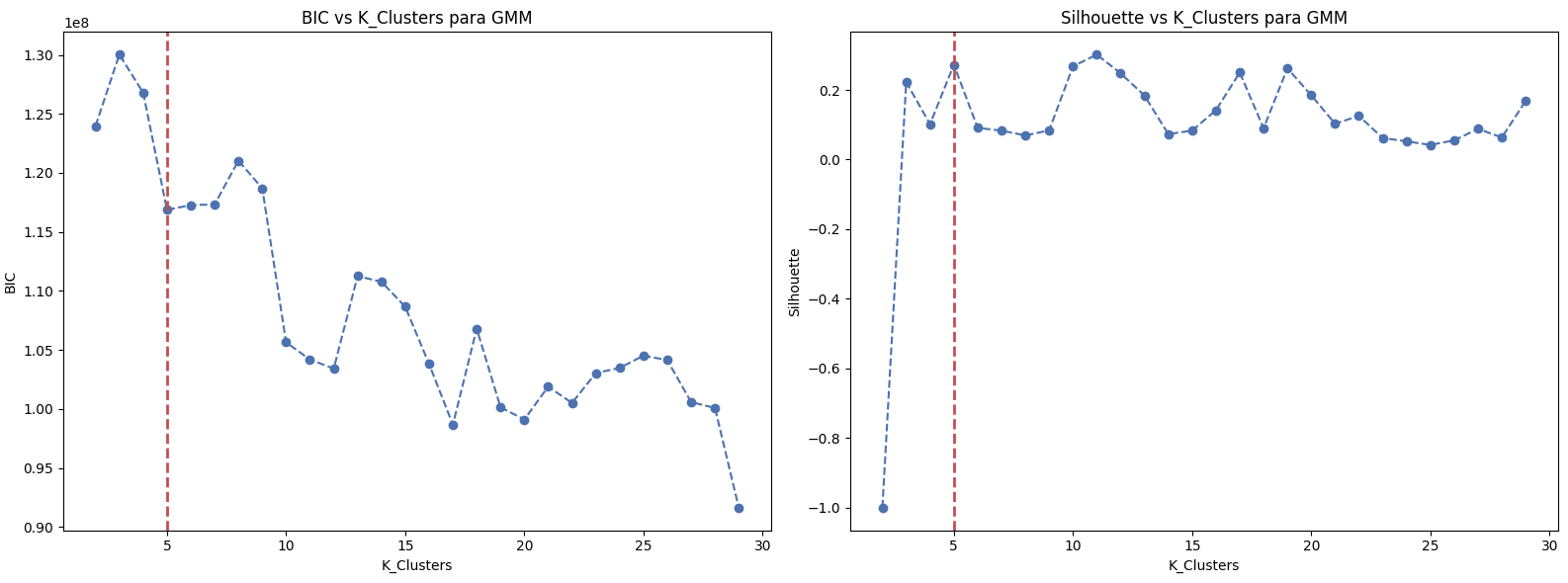}
    \caption{BIC and Silhouette metrics for GMM using Word2Vec embeddings.}
    \label{fig:w2v_clst_mtr}
\end{figure}
As illustrated in Fig.~\ref{fig:w2v_clst_mtr}, the optimal number of clusters lies between $k=3$ and $k=5$, as indicated by consistent elbows in both the BIC and Silhouette curves. Compared to KMeans, GMM provided a better fit to the underlying data distribution, particularly for capturing nuanced semantic substructures.
\begin{figure}[!t]
    \centering
    \includegraphics[width=\columnwidth]{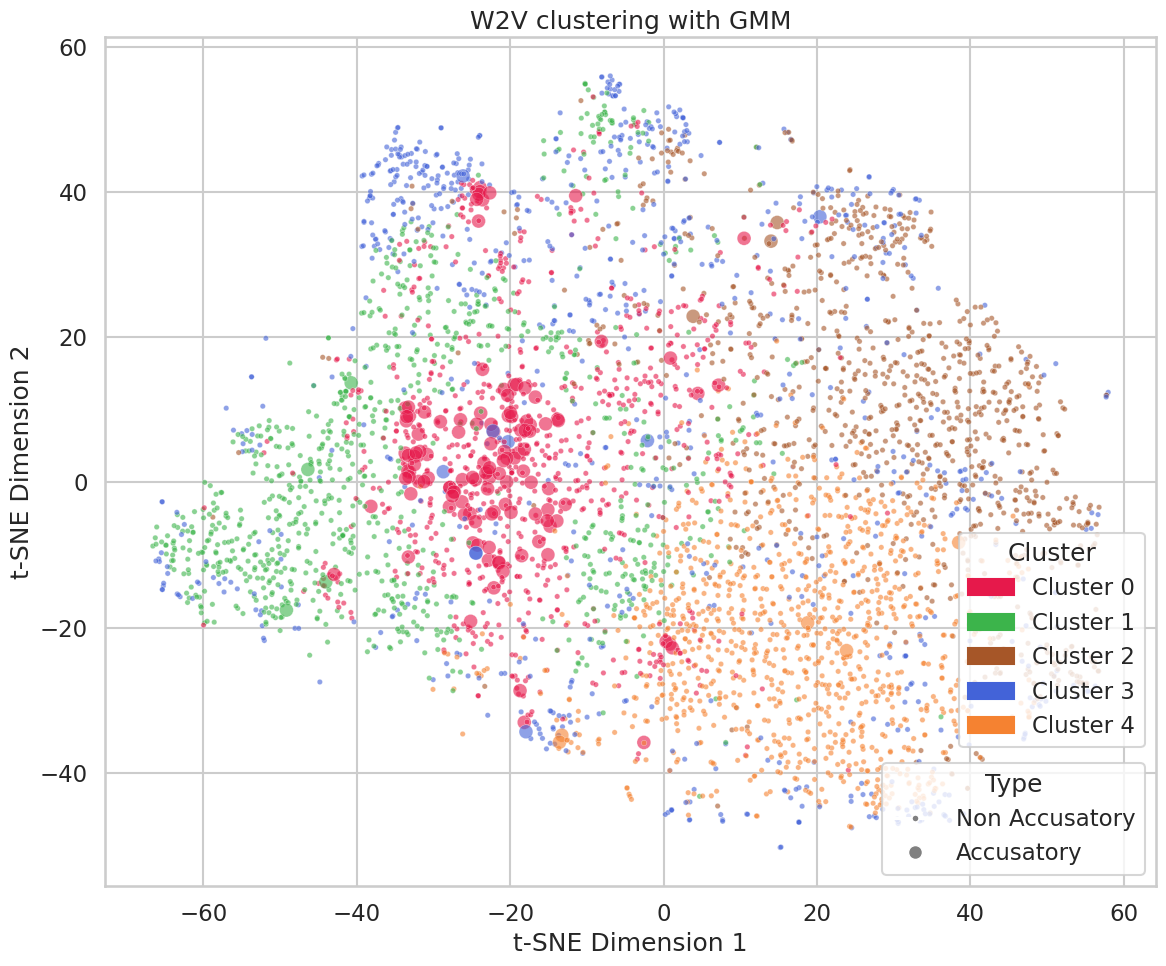}
    \caption{t-SNE projection of the supervised dataset clustered with GMM using Word2Vec embeddings.}
    \label{fig:w2v_clst}
\end{figure}
The t-SNE visualization in Fig.~\ref{fig:w2v_clst} shows that cluster 0 concentrates the highest proportion of accusatory phrases, exhibiting substantially tighter grouping than Transformer-based models.\\

{\textbf{Embedding behavior over clustering:\\}}

{Several studies have evaluated the performance of Transformer-based embeddings across various environments, revealing that their effectiveness often varies depending on the nature of the text. However, the representations derived from models such as LLaMA and RoBERTa can be markedly deficient in similarity analysis tasks, such as clustering} \cite{Comp_LLM_Clustering_2024} \cite{PETUKHOVA2025100} \cite{abs-1908-10084}.

{Under certain conditions, the embeddings generated by the Transformer-based models evaluated in this project suffer from anisotropy. This phenomenon implies that the generated embeddings are confined to a narrow cone within their high-dimensional vector space} \cite{li2020sentenceembeddingspretrainedlanguage}\cite{muennighoff2022sgptgptsentenceembeddings}. {Consequently, semantically distinct texts are mapped to highly similar vector representations, which severely degrades the performance of distance-based clustering algorithms. This limitation stems from the inherent nature of their training objectives; therefore, their suitability for a clustering task remains highly dependent on the specific underlying data.} Table~\ref{tab:cosine_simil_for_cluster} {presents the average cosine similarity within the vector spaces produced by the different algorithms. The data reveals an excessively high average similarity for the RoBERTa and LLaMA vector spaces.}

\begin{table}[htbp]
    \centering
    \caption{ {Average Cosine Similarity by Embedding Technique and Dataset}}
    \label{tab:cosine_simil_for_cluster}
    \begin{tabular}{lccc}
        \hline
        {Dataset Type }& {Word2Vec} & {LLaMA} & {RoBERTa} \\
        \hline
        {Supervised }  & {0.36}     & {0.79}  & {0.91}    \\
        {Unsupervised} & {0.36}     & {0.79}  & {0.91}    \\
        \hline
    \end{tabular}
\end{table}

{In contrast, the space generated by Word2Vec exhibits a significantly lower average cosine similarity. This indicates that Word2Vec embeddings naturally approximate a more isotropic space, where vectors are more uniformly distributed, thereby better capturing the semantic relationships of the data and enabling the definition of clearer and more accurate boundaries between clusters. Beyond anisotropy, the superior performance of Word2Vec in this setting can also be explained by tokenization and domain adaptation factors. Unlike RoBERTa and LLaMA, which rely on native pre-trained tokenizers and representations learned from broad-domain corpora, Word2Vec was trained directly on the Kapak corpus using a custom tokenizer tailored to Spanish public procurement comments. This allowed the model to better preserve domain-specific terminology, informal expressions, spelling variations, and recurrent lexical patterns associated with accusatory language. In contrast, the Transformer-based models were affected by pre-training domain mismatch, since their tokenization and representation spaces were not specifically optimized for noisy, short, and highly specialized procurement-related questions. As a result, even after fine-tuning, their sentence embeddings remained less suitable for distance-based clustering in this particular task.}

{Based on these results, the Word2Vec+GMM configuration was selected for subsequent stages due to its superior clustering performance, semantic alignment, and scalability.}

\subsection{Cluster Identification}
\label{sec:cluster_identification}
Although clustering techniques are effective for revealing latent structures in high-dimensional spaces, they do not directly convey the semantic meaning of each cluster. Consequently, identifying the cluster most representative of accusatory language is critical for enabling robust downstream classification. To this end, a keyword-based identification strategy was adopted, an approach that has proven effective in short-text topic identification under limited supervision \cite{soc_med_clust_comito_2019}. The method evaluates the distribution of predefined accusatory keywords across clusters to identify the most relevant one. Following clustering with the Word2Vec+GMM configuration, the occurrence of five domain-specific keywords—\textit{Direccionado}, \textit{Imposible}, \textit{Ilegal}, \textit{Corrupción}, and \textit{Prohibido}—was analyzed across all clusters. Table~\ref{tab:keywords_cluster}\footnote{Note: Keywords translated from Spanish to English as follows: Direccionado (Directed procurement), Imposible (Impossible), Corrupción (Corruption), Ilegal (Illegal), and Prohibido (Prohibited).} summarizes their distribution in the labeled dataset.

\begin{table}[!t]
\caption{Clusters with Higher Incidence of Key Words – Labeled Dataset}
\label{tab:keywords_cluster}
\small
\setlength{\tabcolsep}{6pt}
\centering
\begin{tabular*}{\columnwidth}{@{}l c c c@{}}
\hline
\makecell{\textbf{Key Word}} & 
\makecell{\textbf{Most Frequent} \\ \textbf{Cluster}} & 
\makecell{\textbf{Frequency}} & 
\makecell{\textbf{Relative} \\ \textbf{Frequency}} \\
\hline
Direccionado & Cluster 0 & 16 & 84.21\% \\
Imposible     & Cluster 0 & 16 & 61.54\% \\
Ilegal        & Cluster 0 & 5  & 83.33\% \\
Corrupción    & Cluster 0 & 4  & 100.00\% \\
Prohibido     & Cluster 0 & 9  & 69.23\% \\
\hline
\end{tabular*}
\end{table}

Table~\ref{tab:keywords_cluster} indicates a strong concentration of all target keywords in Cluster~0, which is consistent with its previously identified high number of accusatory phrases (122 samples). Notably, even the least specific term, \textit{Imposible}, appears in Cluster~0 in more than 61\% of its total occurrences, reinforcing the semantic coherence of this cluster. To assess whether this pattern generalizes beyond labeled data, the same analysis was applied to the unlabeled dataset. The corresponding results are reported in Table~\ref{tab:keywords_cluster_uns}.
\begin{table}[!t]
\caption{Clusters with Higher Incidence of Key Words – Unlabeled Dataset}
\label{tab:keywords_cluster_uns}
\small
\setlength{\tabcolsep}{6pt}
\centering
\begin{tabular*}{\columnwidth}{@{}l c c c@{}}
\hline
\makecell{\textbf{Key Word}} & 
\makecell{\textbf{Most Frequent} \\ \textbf{Cluster}} & 
\makecell{\textbf{Frequency}} & 
\makecell{\textbf{Relative} \\ \textbf{Frequency}} \\
\hline
Direccionado  & Cluster 0 & 198 & 75.00\% \\
Imposible     & Cluster 0 & 358 & 66.42\% \\
Ilegal        & Cluster 0 & 49  & 73.13\% \\
Corrupción    & Cluster 0 & 80  & 94.12\% \\
Prohibido     & Cluster 0 & 151 & 75.50\% \\
\hline
\end{tabular*}
\end{table}
The results in Table~\ref{tab:keywords_cluster_uns} confirm that the dominance of Cluster~0 persists in the unlabeled dataset, supporting the robustness and scalability of the proposed keyword-based cluster identification strategy. Table~\ref{tab:keywords_cluster_uns} confirms that Cluster~0 consistently exhibits the highest concentration of accusatory keywords, with each term occurring more frequently in this cluster than in any other. This consistency supports the reliability of the proposed cluster identification strategy across both labeled and unlabeled datasets. Although Cluster~0 is specific to the current experimental setup and may vary with different initializations or datasets, the keyword-based approach remains independent of cluster indexing, providing a scalable and interpretable mechanism for identifying accusatory content.

\subsection{Classification}
\label{sec:classification}
{Here, to explicitly evaluate the effectiveness of the proposed cascaded pipeline, several baseline and ablation configurations were considered. At the classification stage, GNB and SVM were used as classical supervised learning baselines, while Random Forest was evaluated as the final selected classifier. In addition, two ablation-style scenarios were included: a no-cluster-filter scenario, which evaluates classification over the complete supervised dataset, and a no-SMOTE scenario, which assesses the effect of class imbalance mitigation. Therefore, classification performance was evaluated under two main data scenarios: i) the full supervised dataset without cluster filtering, and ii) the filtered supervised dataset containing only samples from the most accusatory cluster identified through Word2Vec+GMM. Each scenario was further evaluated with and without SMOTE across all classifiers. These baselines allow us to quantify the contribution of both the cluster-based pre-filtering stage and the SMOTE balancing strategy.}

\begin{itemize}
    \item {Full supervised dataset (4841 samples, 143 accusatory phrases).}
    \item {Filtered supervised dataset - only data from most accusatory cluster (962 samples, 122 accusatory phrases).}
\end{itemize}

Three widely used classifiers were evaluated: Random Forest, GNB, and SVM. To further mitigate class imbalance, SMOTE was applied to oversample the minority class, and Stratified K-Fold Cross-Validation was used to ensure balanced training and validation splits. The resulting classification metrics are summarized in Table~\ref{tab:clas_mets}.

\begin{table*}[!t]
\caption{Overall Classification Results With and Without Cluster Filter and SMOTE comparisson}
\label{tab:clas_mets}
\footnotesize
\setlength{\tabcolsep}{6pt} 
\centering
\begin{tabular*}{\textwidth}{@{\extracolsep{\fill}} c c c l l c c c c @{}}
\hline
\multirow{2}{*}{\textbf{Scenario}} & \multirow{2}{*}{\makecell{\textbf{Rows} \\ \textbf{Evaluated}}} & \multirow{2}{*}{\makecell{\textbf{Total No.} \\ \textbf{Acu. Phrases}}} & \multirow{2}{*}{\textbf{Algorithm}} & \multirow{2}{*}{\textbf{Metric}} & \multicolumn{2}{c}{\textbf{with SMOTE}} & \multicolumn{2}{c}{\textbf{without SMOTE}} \\
\cline{6-9}
& & & & & \makecell{\textbf{Negative} \\ \textbf{Class (0)}} & \makecell{\textbf{Positive} \\ \textbf{Class (1)}} & \makecell{\textbf{Negative} \\ \textbf{Class (0)}} & \makecell{\textbf{Positive} \\ \textbf{Class (1)}} \\
\hline
\multirow{9}{*}{\textbf{Cluster Filter}} & \multirow{9}{*}{\textbf{962}} & \multirow{9}{*}{\textbf{122}}
  & \multirow{3}{*}{\textbf{Random Forest}} & Precision & \textbf{0.99} & \textbf{0.84} & 0.98 & 0.91 \\
& & & & Recall    &\textbf{0.97} & \textbf{0.91} & 0.99 & 0.48 \\
& & & & F1-Score  & \textbf{0.98} &\textbf{0.87} & 0.99 & 0.62 \\
\cline{4-9}
& & & \multirow{3}{*}{GNB}           & Precision & 0.96 & 0.41 & 0.99 & 0.33 \\
& & & & Recall    & 0.84 & 0.74 & 0.96 & 0.71 \\
& & & & F1-Score  & 0.9  & 0.52 & 0.97 & 0.46 \\
\cline{4-9}
& & & \multirow{3}{*}{SVM}           & Precision & 0.98 & 0.61 & 0.99 & 0.44 \\
& & & & Recall    & 0.92 & 0.88 & 0.97 & 0.68 \\
& & & & F1-Score  & 0.95 & 0.72 & 0.98 & 0.53 \\
\hline
\multirow{9}{*}{No Cluster Filter} & \multirow{9}{*}{4841} & \multirow{9}{*}{143} 
  & \multirow{3}{*}{Random Forest} & Precision & 1.00    & 0.58 & 0.98 & 1.00    \\
& & & & Recall    & 0.98 & 0.95 & 1.00    & 0.28 \\
& & & & F1-Score  & 0.99 & 0.70  & 0.99 & 0.44 \\
\cline{4-9}
& & & \multirow{3}{*}{GNB}           & Precision & 0.99 & 0.18 & 1.00    & 0.13 \\
& & & & Recall    & 0.88 & 0.83 & 0.81 & 0.90  \\
& & & & F1-Score  & 0.94 & 0.3  & 0.90 & 0.22 \\
\cline{4-9}
& & & \multirow{3}{*}{SVM}           & Precision & 0.99    & 0.38    & 0.99    & 0.28    \\
& & & & Recall    & 0.96    & 0.83    & 0.95    & 0.70    \\
& & & & F1-Score  & 0.98    & 0.52    & 0.97    & 0.40    \\
\hline
\end{tabular*}
\end{table*}

\begin{figure}[!t]
    \centering
    \includegraphics[width=\columnwidth]{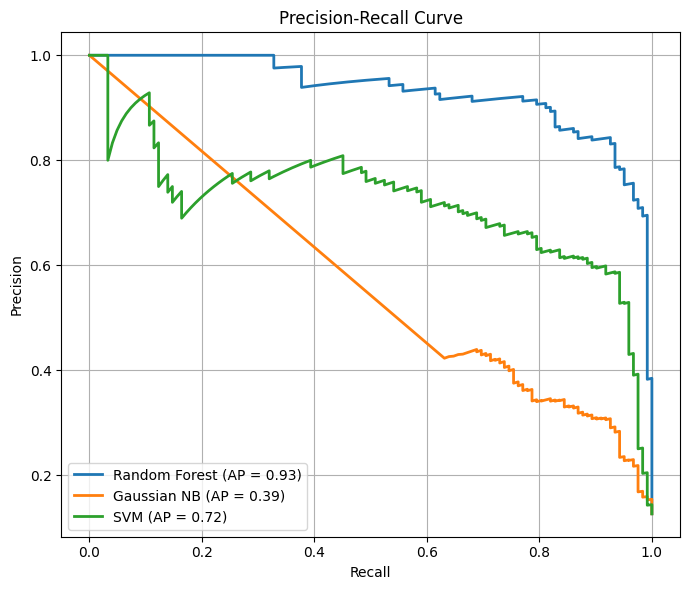}
    \caption{Precision--Recall curve for the Best Classifier Model (Random Forest + SMOTE).}
    \label{fig:pc_Curve}
\end{figure}

Table~\ref{tab:clas_mets} indicates that the Random Forest classifier with SMOTE achieved the best overall performance, maintaining a strong balance between precision and recall and yielding an F1-score of 0.87 for the positive (accusatory) class. The Precision–Recall curve in Fig.~\ref{fig:pc_Curve} further demonstrates the robustness of the Random Forest model across decision thresholds. An Average Precision (AP) score of 0.93 confirms its effectiveness under severe class imbalance. 

{On the other hand, as anticipated, the SMOTE strategy significantly improved classification performance across all classifiers. Furthermore, the filtering strategy based on the most accusatory cluster identified via Word2Vec+GMM proved to be extremely robust compared to the other scenarios. It was observed that reducing data imbalance through cluster-based filtering not only yielded more efficient results but also considerably decreased execution time due to the substantial reduction in the training set size.}

\begin{figure}[!t]
    \centering
    \includegraphics[width=\columnwidth]{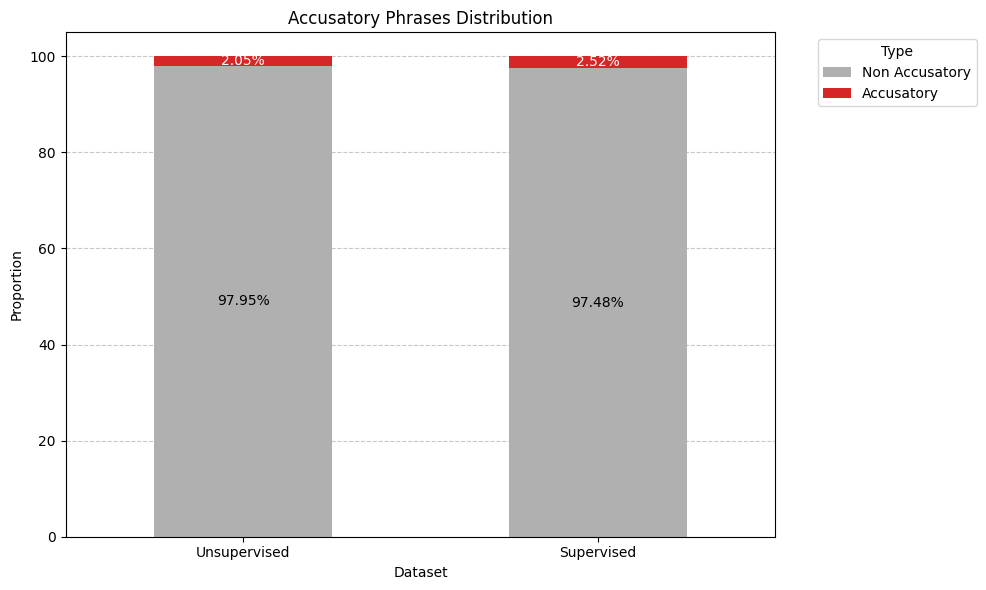}
    \caption{Proportion of accusatory phrases in the most accusatory cluster relative to the full dataset.}
    \label{fig:dist_datasets_res}
\end{figure}

Finally, Fig.~\ref{fig:dist_datasets_res} shows the proportion of accusatory phrases contained in the most accusatory cluster relative to the total size of each dataset, for both supervised and unsupervised cases. In both distributions, non-accusatory samples account for more than 97\% of the data, highlighting the ability of the clustering strategy to effectively isolate semantically accusatory content. The near-identical proportions across datasets indicate strong semantic alignment between clustering and classification objectives, further validating the consistency of the proposed pipeline when applied to unlabeled data.

\subsection{Unsupervised Testing}
\label{sec:unsupervised_testing}
After selecting the Random Forest + SMOTE classifier as the best-performing model in the supervised phase, the pipeline was extended to the unlabeled dataset to assess its generalization capability in the absence of ground-truth labels. Following the previously described methodology, the unsupervised evaluation consisted of two main steps: $i)$ filtering the unlabeled dataset to retain only records belonging to the most accusatory cluster identified via Word2Vec+GMM, and $ii)$ applying the pre-trained Random Forest classifier to the filtered subset to generate predictions.
\begin{table}[!t]
\caption{Final Distribution of Accusatory Samples and Model-Identified Candidates}
\label{tab:acu_cluster_summary}
\small
\setlength{\tabcolsep}{4pt}
\centering
\begin{tabular*}{\columnwidth}{@{}l c c c@{}}
\hline
\textbf{Dataset} & 
\makecell{\textbf{Total Rows}} & 
\makecell{\textbf{Real Accu.} \\ \textbf{Phrases}} & 
\makecell{\textbf{Accu. Phrases in} \\ \textbf{Most Accu.} \\ \textbf{Cluster}} \\
\hline
Unsupervised & 92{,}579 & --  & 1{,}892 \\
Supervised   & 4{,}841  & 143 & 122 \\
\hline
\end{tabular*}
\end{table}
{A summary of the resulting predictions is reported in } Table~\ref{tab:acu_cluster_summary}, {which compares the distribution of accusatory samples or candidates between the supervised and unsupervised datasets. Since the unlabeled dataset does not contain ground-truth annotations, the reported value for this dataset corresponds to model-identified accusatory candidates obtained after applying the Word2Vec+GMM filtering stage and the trained Random Forest classifier. In the supervised dataset, 122 out of 4,841 samples were retained as accusatory within the most accusatory cluster, while in the unlabeled dataset, 1,892 out of 92,579 samples were identified as accusatory candidates. These proportions are of the same order of magnitude, suggesting that the model behaves consistently when extended to unlabeled data and supporting the reliability of the clustering stage as a pre-filtering mechanism. To further assess the reliability of the model beyond the controlled supervised evaluation, a manual validation step was conducted over a random subset of positive predictions obtained from the unlabeled dataset. This review aimed to estimate the real-world precision of the proposed pipeline when applied to previously unseen procurement comments. The manual inspection confirmed that most predicted accusatory samples corresponded to comments explicitly suggesting corruption, bid manipulation, directed procurement, illegality, or procedural irregularity. The observed empirical precision was approximately 0.84, which is consistent with the supervised evaluation and supports the practical usefulness of the proposed model in real-world monitoring scenarios.}

{These results also reveal an important limitation of the current pipeline. Indirect, ironic, or sarcastic accusations may be difficult to detect because they often lack explicit lexical markers of corruption, bid manipulation, or procedural irregularity. Similarly, strongly negative comments may resemble accusatory language even when they do not contain a clear allegation. Future work should address these failure modes by expanding the annotation scheme to include indirect and sarcastic accusations, incorporating process-level contextual information, and exploring contrastive or sentence-level Transformer architectures specifically optimized for nuanced semantic similarity. Human-in-the-loop validation could also be incorporated to review ambiguous cases and progressively enrich the labeled dataset.}
\subsection{Final Performance Metrics}
\label{sec:fin_perf_met}
The proposed workflow involves multiple training and processing stages with varying computational demands. To evaluate its efficiency and reproducibility, a performance assessment was conducted across the main stages of the pipeline. The following considerations summarize the most relevant aspects.
\begin{itemize}
    \item \textbf{Word Embedding Generation.} Embeddings were generated for both supervised and unsupervised datasets. For RoBERTa and LLaMA, both base and fine-tuned models were evaluated.
    \item \textbf{Clustering.} K-Means and Gaussian Mixture Models were applied to assess clustering efficiency and stability.
    \item \textbf{Classification.} Performance was evaluated using Random Forest, GNB, and SVM classifiers. The identification of the most accusatory cluster was performed prior to classification.
\end{itemize}
Overall, the complete pipeline based on Word2Vec required approximately 157 minutes for full training. This configuration demonstrated conservative CPU and RAM usage and did not require GPU resources, making it suitable for environments with limited computational infrastructure.
\subsubsection{Near Real-Time Application}
Experimental results indicate that, when using pre-trained models, predicting whether an unlabeled sentence is accusatory requires approximately 1 to 3 seconds. Consequently, the Word2Vec+GMM+Random Forest pipeline is suitable for near real-time deployment. Model enrichment through retraining can be performed efficiently when leveraging the optimal parameters identified in this study. Training times remain modest, as summarized in Table~\ref{tab:training_times} for the best-performing configuration.
\begin{table}[!t]
\caption{Training Time for Best Parameters (W2V + GMM + Random Forest)}
\label{tab:training_times}
\small
\setlength{\tabcolsep}{10pt}
\centering
\begin{tabular*}{20.25pc}{@{}l c c@{}}
\hline
\textbf{Step} & \textbf{Dataset Size} & \textbf{Time} \\ 
\hline
W2V Training & 92{,}579 & 3 min \\
W2V Embedding Generation & 92{,}579 & 4 min \\
GMM Clustering ($k=5$) & 92{,}579 & 3 min \\
Random Forest Classification & 4{,}841 & 1 min \\ 
\hline
\multicolumn{2}{l}{\textbf{Total}} & 11 min \\
\hline
\end{tabular*}
\end{table}
These results indicate that adapting the pipeline to new domains or enriching it with additional data can be achieved efficiently, with low computational cost and short retraining times.

\section{Cross-Country Generalizability and Data Availability in Latin America}
\label{sec:generalizability}
Although the proposed NLP pipeline was developed using data from Ecuador's SOCE platform, an important question is whether the methodology can be generalized to other public procurement systems in Latin America. To explore this, a comparative review of major procurement portals in the region was conducted, focusing on: $(i)$ the existence of clarification or Q\&A mechanisms during the pre-contractual stage, $(ii)$ the availability of these interactions in machine-readable formats, and $(iii)$ the presence of open data portals supporting bulk access. Overall, most countries publish supplier questions or clarification records as part of their procurement processes. However, in the majority of cases, these interactions are available only as embedded HTML pages or PDF documents intended for human consumption, with limited or no support for large-scale automated extraction. In several jurisdictions, additional legal or technical barriers further restrict web scraping or bulk access to Q\&A data. For instance, Colombia’s SECOP II provides rich interaction content but does not allow exporting complete Q\&A histories through its open data portal. Similar limitations are observed in Argentina, Mexico, Brazil, and Peru. By contrast, Ecuador’s SOCE platform stands out as one of the few systems in the region that systematically records Q\&A interactions as structured, time-stamped HTML components, without explicit regulatory restrictions on automated extraction. This unique combination of accessibility and structure made Ecuador an appropriate pilot case for this study. Nevertheless, the proposed pipeline is model-agnostic and embedding-based, and can be readily adapted to other national contexts provided that structured or semi-structured textual data becomes available.
Table~\ref{tab:latam-minimal} summarizes a representative subset of major Latin American procurement systems, highlighting the practical constraints and opportunities for extending the proposed approach beyond Ecuador.
{\scriptsize
\begin{table*}[!t]
\centering
\caption{Availability of Clarifications, Official Portals, and Open Data in Major Latin American Procurement Systems (2019--2024)}
\label{tab:latam-minimal}
\renewcommand{\arraystretch}{1.2}
\begin{tabular}{p{2.4cm} p{3.2cm} p{3.4cm} p{4.8cm}}
\hline
\textbf{Country} &
\textbf{Official Portal} &
\textbf{Clarification Q\&A Mechanism?} &
\textbf{How to Access Q\&A Data} \\
\hline

Chile &
Mercado Público (ChileCompra) &
Yes – Clarification Forum &
Exportable per tender via Excel; structured and accessible. \\

Colombia &
SECOP II – Colombia Compra Eficiente &
Yes – Observations Module &
Accessible per process; no bulk or machine-readable Q\&A export. \\

Brazil &
Compras.gov.br / Portal Nacional de Contratações Públicas (PNCP) &
Yes – Esclarecimentos &
Published as documents; no dedicated Q\&A dataset. \\

Mexico &
CompraNet &
Yes – Clarification Meetings &
Clarification acts downloadable as PDF documents. \\

Argentina &
COMPR.AR &
Yes – Consultations and Clarifications &
Available only as PDFs; unstructured. \\

Peru &
SEACE – OSCE &
Yes – Consultations and Observations &
Consultation acts downloadable as PDFs. \\

Ecuador &
SOCE – SERCOP &
Yes – Aclaraciones (structured) &
Structured HTML; fully extractable at scale. \\

Costa Rica &
SICOP &
Yes – Aclaraciones &
Open Data portal allows CSV/Excel download. \\

Uruguay &
ARCE – SICE &
Yes – Inquiry Reports &
Reports downloadable; Q\&A text not included in Open Contracting Data Standard (OCDS). \\

Panama &
PanamaCompra &
Yes – Homologation Consultations &
Clarification acts included in tender documents (PDF). \\

Dominican Republic &
DGCP – Transactional Portal (SECP) &
Yes – Communications and Meetings &
Acts downloadable from process files; no Q\&A dataset. \\

Guatemala &
GUATECOMPRAS &
Yes – Clarification Meetings &
Clarification acts downloadable as PDFs; unstructured. \\

\hline
\end{tabular}
\end{table*}
\textit{Sources: official national public-procurement and open-data portals of the corresponding countries, accessed in 2026.}
}
In summary, while the proposed methodological framework is broadly generalizable, its practical deployment across Latin America is currently constrained primarily by data accessibility rather than by modeling limitations. These findings motivate future collaboration with regional open contracting and transparency initiatives to promote standardized, machine-readable publication of supplier–buyer interactions.

{From} Table~\ref{tab:latam-minimal}, {although Chile and Costa Rica provide comparatively more structured access to procurement-related information, a direct zero-shot evaluation of the proposed pipeline on these countries was not included in the present study. Such an experiment would require constructing a comparable out-of-distribution dataset, including data extraction, normalization, and manual validation under the same operational definition of accusatory language used for the SOCE corpus. In addition, differences in procurement terminology, portal structure, and country-specific administrative language may affect both embedding generation and cluster identification. Therefore, while the proposed pipeline is portable at the methodological level, its empirical validation in other national contexts requires country-specific dataset construction and annotation.}

\section{Conclusions and Future Research Directions}
\label{sec:conclu}
This study highlighted the importance of strategic customization in NLP workflows for real-world governance and auditing applications. By regularizing textual data and employing a domain-specific tokenizer, a comparatively simple embedding model such as Word2Vec was shown to outperform more complex alternatives, demonstrating that effective semantic extraction did not necessarily require resource-intensive architectures. This result supported the development of efficient, scalable, and accessible ML/NLP systems suitable for environments with limited computational infrastructure. The proposed pipeline further demonstrated that keyword-based cluster identification constituted a robust and cost-effective strategy when dealing with sentiment-laden or strongly opinionated language, as commonly observed in procurement-related complaints and allegations. The integration of supervised and unsupervised learning stages was shown to be essential for improving both performance and generalizability. In particular, the combination of Word2Vec embeddings, GMM-based clustering, and a Random Forest classifier yielded strong classification results under severe class imbalance, provided that appropriate validation strategies (i.e., SMOTE, Stratified K-Fold cross-validation, and Average Precision metrics) were applied. These findings emphasized the importance of balanced evaluation practices, as reliance on classical metrics alone could have led to misleading conclusions in applied scenarios.
Overall, this work demonstrated that transparent and effective NLP pipelines for corruption risk detection could be built with modest computational cost. Beyond public procurement auditing, the proposed approach was found to be applicable to related domains such as user recommendation systems or social networks, where semantic relationships, informal language, and noisy text were prevalent. Additionally, the implemented pipeline proved to be tolerant to spelling mistakes, grammatical inconsistencies, and syntactic variations, further supporting its robustness.

Based on the results obtained, several directions were identified for future research: $i)$ a fully operational semi-supervised workflow could be developed, enabling continuous model updates at predefined time intervals as new data becomes available; $ii)$ further comparative analyses of word embedding models could be conducted by incorporating additional architectures such as ALBERT, T5, and more recent or larger LLaMA variants, particularly in combination with clustering-based pre-filtering strategies; $iii)$ alternative unsupervised learning techniques, including HDBSCAN and hierarchical clustering, could be explored to further improve the identification of semantically coherent and accusatory clusters. 
{$iv)$ Although the initial embeddings generated by RoBERTa and LLaMA did not yield the best results for this specific clustering task, their underlying potential remains significant. Future work should therefore not discard these architectures, but rather strengthen their adaptation strategies. In particular, continuous fine-tuning with larger domain-specific corpora, contrastive learning objectives, or sentence-embedding architectures such as Siamese BERT-Networks (Sentence-BERT) could substantially improve their performance. Similarly, recent frameworks such as LLM2Vec, which employ multi-stage adaptation strategies to improve the isotropy and semantic quality of LLM-based embeddings, represent a promising direction for future research.}
{v) Future work should conduct a cross-country empirical validation of the proposed pipeline using structured procurement data from countries such as Chile or Costa Rica. This would allow the evaluation of zero-shot, lightly fine-tuned, and country-adapted configurations, providing stronger evidence of the regional portability of the methodology.}
{vi) Another important future direction involves moving from the current modular cascaded pipeline toward partially joint or end-to-end optimization strategies. In this study, clustering and classification were intentionally decoupled to preserve interpretability, modularity, and independent validation under limited labeled data. However, future work could explore semi-supervised architectures in which cluster assignments and classification objectives are optimized jointly, while incorporating regularization mechanisms to avoid overfitting and preserve the semantic interpretability of the discovered clusters.}
\cite{abs-1908-10084} \cite{behnamghader2024llm2veclargelanguagemodels}.


 \section*{Acknowledgments}
 The authors gratefully acknowledge the Sigma AI Lab at Universidad San Francisco de Quito (USFQ) for providing the computational resources and AI infrastructure that enabled this research, and the Deutsche Gesellschaft für Internationale Zusammenarbeit (GIZ) GmbH for its support in the development of Kapak.



%

\bibliographystyle{ieeetr}
\bibliography{bibfile}

\vfill

\end{document}